\pdfoutput=1

\documentclass[11pt]{article}
\usepackage[]{ACL2023}

\usepackage{amsmath,amsfonts,bm}

\def\eqref#1{equation~\ref{#1}}

\def\1{\bm{1}}

\DeclareMathAlphabet{\mathsfit}{\encodingdefault}{\sfdefault}{m}{sl}
\SetMathAlphabet{\mathsfit}{bold}{\encodingdefault}{\sfdefault}{bx}{n}

\usepackage{times}
\usepackage{latexsym}
\usepackage[T1]{fontenc}
\usepackage{microtype}
\usepackage{inconsolata}
\usepackage{stfloats}
\usepackage{times}
\usepackage{soul}
\usepackage{url}
\usepackage{graphicx}
\usepackage{amsmath}
\usepackage{amsthm}
\usepackage{algorithm}
\usepackage{algorithmic}
\usepackage[switch]{lineno}
\usepackage{latexsym}
\usepackage{microtype}
\usepackage{inconsolata}
\usepackage{booktabs}
\usepackage{enumitem}
\usepackage{amsfonts}
\usepackage{float}
\usepackage[noabbrev,capitalize]{cleveref}

\title{History Semantic Graph Enhanced Conversational KBQA \\ with Temporal Information Modeling}

\author{
Hao Sun\textsuperscript{\rm 1},
Yang Li\textsuperscript{\rm 2},
Liwei Deng\textsuperscript{\rm 3},
Bowen Li\textsuperscript{\rm 2},
Binyuan Hui\textsuperscript{\rm 2}\\
\textbf{
Binhua Li\textsuperscript{\rm 2},
Yunshi Lan\textsuperscript{\rm 4},
Yan Zhang\textsuperscript{\rm 1},
Yongbin Li \textsuperscript{\rm 2}
}
\\
\textsuperscript{\rm 1} 
Peking University,
\textsuperscript{\rm 2}
Alibaba Group \\
\textsuperscript{\rm 3}
University of Electronic Science and Technology of China,
\textsuperscript{\rm 4}
East China Normal University
\\
\small \tt{sunhao@stu.pku.edu.cn}, \tt{zhyzhy001@pku.edu.cn}\\
\small \tt{\{ly200170, binyuan.hby, binhua.lbh, shuide.lyb\}@alibaba-inc.com} \\
\small \tt{deng\_liwei@std.uestc.edu.cn}, \tt{libowen.ne@gmail.com}, \tt{yslan@dase.ecnu.edu.cn} \\
}

\begin{document}
\maketitle
\begin{abstract}
Context information modeling is an important task in conversational KBQA. However, existing methods usually assume the independence of utterances and model them in isolation. In this paper, we propose a \textbf{H}istory \textbf{S}emantic \textbf{G}raph \textbf{E}nhanced KBQA model (\textbf{HSGE}) that is able to effectively model long-range semantic dependencies in conversation history while maintaining low computational cost. The framework incorporates a context-aware encoder, which employs a dynamic memory decay mechanism and models context at different levels of granularity. We evaluate HSGE on a widely used benchmark dataset for complex sequential question answering. Experimental results demonstrate that it outperforms existing baselines averaged on all question types.

\end{abstract}

\section{Introduction}
In recent years, with the development of large-scale knowledge base (KB) like DBPedia~\cite{auer2007dbpedia} and Freebase~\cite{bollacker2008freebase}, Knowledge Base Question Answering~(KBQA) \cite{wang2020modelling,ye2021rng,yan2021large,yadati2021knowledge,das2021case,wang2022new} has become a popular research topic, which aims to convert a natural language question to a query over a knowledge graph to retrieve the correct answer. With the increasing popularity of AI-driven assistants (e.g., Siri, Alexa and Cortana), research focus has shifted towards conversational KBQA \cite{shen2019multi,kacupaj2021conversational,marion2021structured} that involves multi-turn dialogues.

\begin{figure}[ht]
	\centering
\includegraphics[width=0.47\textwidth]{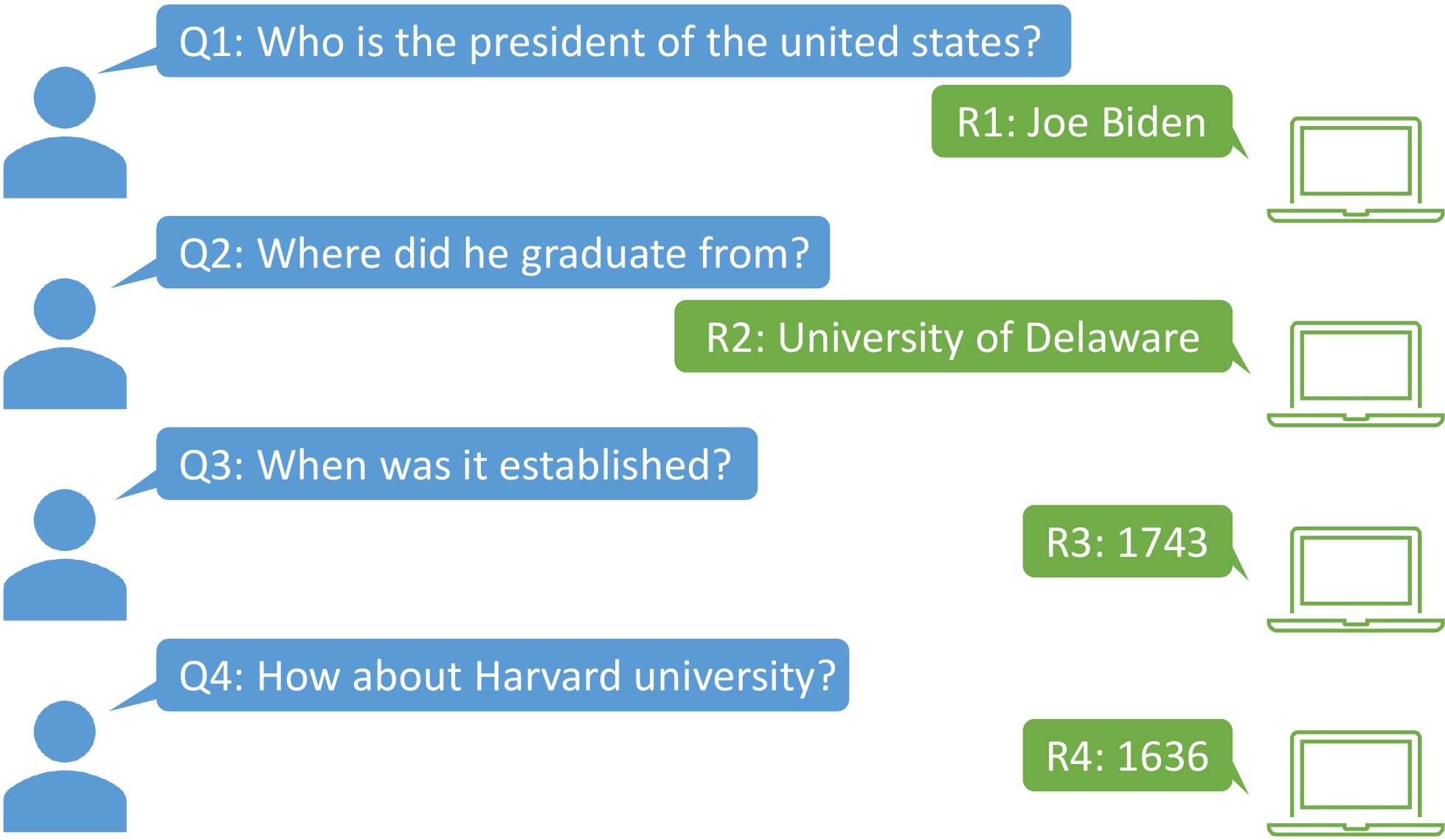}
	\caption{An example illustrating the task of conversational KBQA.}
	\label{image:running_example}
\end{figure}

A common solution to the task of conversational KBQA is to map an utterance to a logical form using semantic parsing approach~\cite{shen2019multi,guo2018dialog}. 
The state-of-the-art semantic parsing approach~\cite{kacupaj2021conversational}  breaks down the process into two stages: a logical form is first generated by low-level features and then the missing details are filled by taking both the question and templates into consideration. Other approaches~\cite{dong2016language,liang2016neural,guo2018dialog} mainly focus on first detecting entities in the question and then mapping the question to a logical form. 

Despite the inspiring results of the semantic parsing methods mentioned above, most of them fail to model the long-range semantic dependency in conversation history. Specifically, they usually directly incorporate immediate two turns of conversations and ignore the conversation history two turns away. To demonstrate the importance of long-range conversation history, \cref{image:running_example} shows an example illustrating the task of conversational KBQA. After the question ``who is the president of the United States'', the user consecutively proposes three questions that involve \texttt{Coreference} and \texttt{Ellipsis} phenomena~\cite{androutsopoulos1995natural}. Only when the system understands the complete conversation history can the system successfully predict the answer. Though existing contextual semantic parsing models~\cite{iyyer-etal-2017-search,suhr-etal-2018-learning,yu-etal-2019-cosql} can be used to model conversation history, a survey~\cite{liu2020far} points out that their performance is not as good as simply concatenating the conversation history, which is the most common conversation history modeling technique.

To tackle the issues mentioned above, we propose a \textbf{H}istory \textbf{S}emantic \textbf{G}raph \textbf{E}nhanced Conversational KBQA model (HSGE) for conversation history modeling. 
Specifically, we convert the logical forms of previous turns into history semantic graphs, whose nodes are the entities mentioned in the conversation history and edges are the relations between them. By applying graph neural network on the history semantic graph, the model can capture the complex interaction between the entities and improve its understanding of the conversation history. From the perspective of practice, using the history semantic graph to represent the conversation history is also more computationally efficient than directly concatenating the conversation history.
Besides, we design a context-aware encoder that addresses user's  conversation focus shift phenomenon~\cite{lan2021modeling} by introducing temporal embedding and allows the model to incorporate information from the history semantic graph at both token-level and utterance-level.

To summarize, our major contributions are:

\begin{itemize}
\item We propose to model conversation history using history semantic graph, which is effective and efficient. As far as we know, this is the first attempt to use graph structure to model conversation history in conversational KBQA.
\item We design a context-aware encoder that utilizes temporal embedding to address the shift of user's conversation focus and aggregate context information at different granularities.
\item Extensive experiments on the widely used CSQA dataset demonstrate that HSGE achieves the state-of-the-art performance averaged on all question types.
\end{itemize}

\section{Related Work}
The works most related to ours are those investigating semantic parsing-based approaches in conversational KBQA. Given a natural language question, traditional semantic-parsing methods \cite{zettlemoyer2009learning,artzi2013weakly} usually learn a lexicon-based parser and a scoring function to produce a logical form. For instance, \cite{zettlemoyer2009learning} propose to learn a context-independent CCG parser and \cite{long2016simpler} utilizes a shift-reduce parser for logical form construction.

Recently, neural semantic parsing approaches are gaining attention with the development of deep learning~\cite{qu2019attentive, chen2019graphflow}. For example, \cite{liang2016neural} introduces a neural symbolic machine (NSM) extended with a key-value memory network.
\cite{guo2018dialog} proposes D2A, a neural symbolic model with memory augmentation. S2A+MAML~\cite{guo2019coupling} extends D2A with a meta-learning strategy to account for context.
\cite{shen2019multi} proposes the first multi-task learning framework MaSP that simultaneously learns type-aware entity detection and pointer-equipped logical form generation.
\cite{plepi2021context} introduces CARTON which utilizes pointer networks to specify the KG items.
\cite{kacupaj2021conversational} proposes a graph attention network to exploit correlations between entity types and predicates.
\cite{marion2021structured} proposes to use KG contextual data for semantic augmentation.

While these methods have demonstrated promising results, they typically only consider the immediate two turns of conversations as input while neglecting the context two turns away. 
Though \cite{guo2018dialog} introduces a Dialog Memory to maintain previously observed entities and predicates, it fails to capture their high-order interaction information.
By introducing history semantic graph, our model HSGE can not only memorize previously appeared entities and predicates but also model their interaction features using GNN to gain a deeper understanding of conversation history.

\section{Method}

\begin{figure*}[ht]
	\centering
	\includegraphics[width=\textwidth]{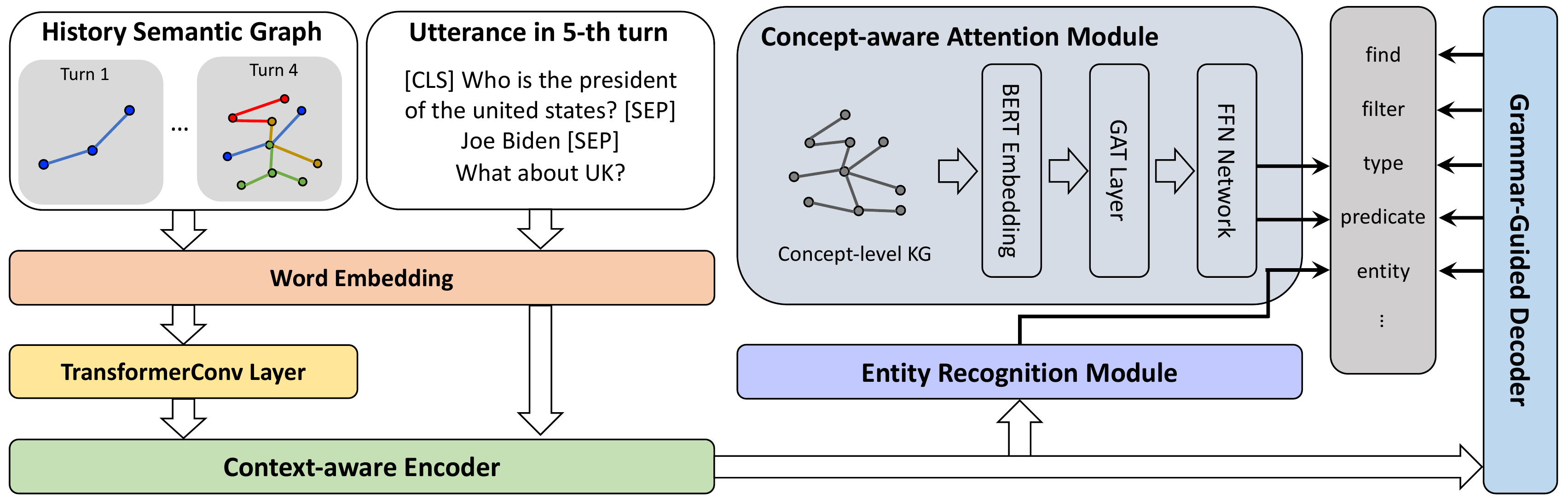}
	\caption{Model architecture of HSGE, which includes Word Embedding, TransformerConv Layer, Context-aware Encoder, Entity Recognition Module, Concept-aware Attention Module and Grammar-Guided Decoder.}
	\label{image:model}
\end{figure*}

The structure of our proposed HSGE model is illustrated in \cref{image:model}. The model consists of six components: 
Word Embedding, TransformerConv Layer, Context-aware Encoder, Entity Recognition Module, Concept-aware Attention Module
and Grammar-Guided Decoder.

\subsection{Grammar}
We predefined a grammar with various actions in \cref{table:grammar}, which can result in different logical forms that can be executed on the KG. Analogous to \cite{kacupaj2021conversational}, each action in this work consists of three components: a semantic category, a function symbol and a list of arguments with specified semantic categories. Amongst them, semantic categories can be classified into two groups depending on the ways of instantiation. One is referred to as entry semantic category (i.e., $\{e, p, tp, num\}$ for entities, predicates, entity types and numbers) whose instantiations are constants parsed from a question. Another is referred to as intermediate semantic category (i.e., $\{ set, dict, boolean, number\}$) whose instantiation is the output of an action execution.

\subsection{Input and Word Embedding}
To incorporate the recent dialog history from previous interactions, the model input for each turn contains the following utterances: the previous question, the previous answer and the current question. Utterances are separated by a \texttt{[SEP]} token and a context token \texttt{[CLS]} is appended at the beginning of the input as the semantic representation of the entire input.

Specifically, given an input $u$, we use WordPiece tokenization~\cite{wu2016google} to tokenize the conversation context into token sequence $\{w_1, ...,w_n \}$, and then we use the pre-trained language model BERT~\cite{devlin2018bert} to embed each word into a vector representation space of dimension $d$. Our word embedding module provides us with an embedding sequence $\{x_1,..., x_n \}$, where $x_i \in \mathbb{R}^d$ is given by $x_i = \texttt{BERT}(w_i)$.

\subsection{History Semantic Graph}

\begin{figure}[ht]
	\centering
	\includegraphics[width=0.48\textwidth]{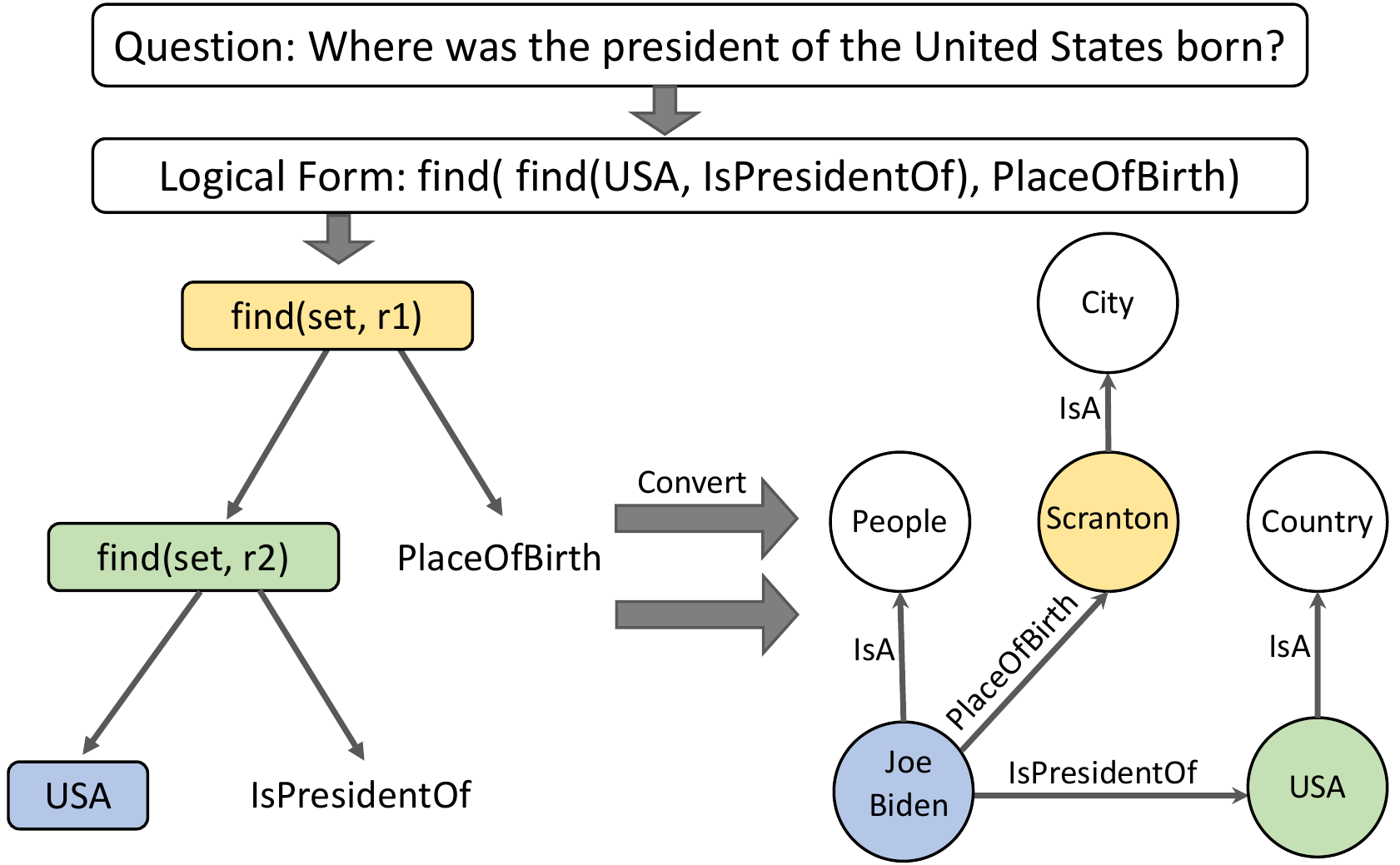}
	\caption{Illustration example for history semantic graph construction.}
	\label{image:graph}
\end{figure}

To effectively and efficiently model conversation history that contains multiple turns, we design \textbf{History Semantic Graph}, inspired by the recent studies on dynamically evolving structures~\cite{hui2021dynamic}.
As the conversation proceeds, more and more entities and predicates are involved, which makes it difficult for the model to capture the complex interactions among them and reason over them. Thus, we hope to store these information into a graph structure and empower the model with strong reasoning ability by applying GNN onto the graph.
Considering that we are trying to model the interactions between entities and predicates which are naturally included in logical forms, one good solution is to directly convert the logical forms into KG triplets as shown in \cref{image:graph}. By doing so, we guarantee the quality of the graph because the entities and predicates are directly related to the answers of previous questions, while also injecting history semantic information into the graph.

\paragraph{Graph Construction.}

Specifically, we define the history semantic graph to be $\mathcal{G} =  <\mathcal{V}, \mathcal{E}>$, where $\mathcal{V} = set(e) \cup set(tp)$, $\mathcal{E} = set(p)$, and $e, tp, p$ denote entity, entity type and predicate, respectively.
We define the following rules to transform the actions defined in Table~\ref{table:grammar} to the KG triplets:

\begin{itemize}
\item For each element $e_i$ in the operator result of $set\rightarrow find(e,p)$, we directly add  <$e_i, p, e$> into the graph.
\item For each element $e_i$ in the operator result of $set\rightarrow find\_reverse(e,p)$, we directly add <$e, p, e_i$> into the graph.
\item For each entity $e_i \in \mathcal{V}$, we also add the <$e_i, IsA, tp_i$> to the graph, where $tp_i$ is the entity type of entity $e_i$ extracted from Wikidata knowledge graph.
\item For the $find$ and $find\_reverse$ actions that are followed by $filter\_type$ or $filter\_multi\_types$ action for entity filtering, we would add the element in the filtering result to the graph, which prevents introducing unrelated entities into the graph.
\end{itemize}

It is worth mentioning that we choose to transform these actions because they directly model the relationship between entities and predicates. Besides, as the conversation proceeds and new logical forms are generated, more KG triplets will be added to the graph and the graph will grow larger.  However, the number of nodes involved in the graph is still relatively small and is highly controllable by only keeping several recent KG triplets. Considering the $O(N^2)$ computational complexity of Transformer encoders~\cite{vaswani2017attention}, it would be more computationally efficient to model conversation history using history semantic graph than directly concatenating previous utterances.

\paragraph{Graph Reasoning.}

Given constructed history semantic graph $\mathcal{G}$, we first initialize the embeddings of nodes and relations using BERT, i.e., $\texttt{BERT}(e_i/p_i)$, where $e_i$ and $p_i$ represent the text of node and relation, respectively. Then we follow TransformerConv~\cite{shi2020masked} and update node embeddings as follows:

\begin{equation}
H = \textrm{TransformerConv}(E, \mathcal{G})
\end{equation}

where $E \in \mathbb{R}^{(|\mathcal{V}| +|\mathcal{E}|) \times d}$ denotes the embeddings of nodes and relations.

\subsection{Context-aware Encoder}
\paragraph{Temporal Information Modeling.}
As the conversation continues and further inquiries are raised, individuals tend to focus more on recent entities, which is also called \texttt{Focal Entity Transition} phenomenon~\cite{lan2021modeling}.
To incorporate this insight into the model, we introduce temporal embedding to enable the model to distinguish newly introduced entities. 
Specifically, given the current turn index $t$ and previous turn index $i$ in which entities appeared, we define two distance calculation methods:
\begin{itemize}
    \item \textbf{Absolute Distance}: The turn index of the previous turn in which the entities were mentioned, i.e., $D = t$.
    \item \textbf{Relative Distance}: The difference in turn indices between the current turn and the previous turn in which the entities were mentioned, i.e., $D = t - i$.
\end{itemize}

For each method, we consider two approaches for representing the distance: unlearnable positional embedding and learnable positional embedding. For unlearnable positional encoding, the computation is defined using the following sinusoid function~\cite{vaswani2017attention}:
\begin{equation}
\left\{ \begin{array}{l}
e_t(2i) = sin(D/10000^{2i/d}), \\ 
e_t(2i+1) = cos(D/10000^{2i/d}), 
\end{array}\right.
\end{equation}
where $i$ is the dimension and $D$ is the absolute distance or relative distance. 

For learnable positional encoding, the positional encoding is defined as a learnable matrix $E_t \in \mathbb{R}^{M \times d} $, where $M$ is  the predefined maximum number of turns.

Then we directly add the temporal embedding to obtain temporal-aware node embeddings.
\begin{equation}
    \bar{h}_i = h_i + e_t,
\end{equation}
where $h_i$ is the embedding of node $e_i$.

\paragraph{Semantic Information Aggregation.}
As the conversation progresses, user's intentions may change frequently, which leads to the appearance of intention-unrelated entities in history semantic graph. To address this issue, we introduce token-level and utterance-level aggregation mechanisms that allow the model to dynamically select the most relevant entities. These mechanisms also enable the model to model contextual information at different levels of granularity.

\begin{itemize}
    \item \textbf{Token-level Aggregation}: 
    For each token $x_i$, we propose to attend all the nodes in the history semantic graph to achieve fine-grained modeling at token-level:
    \begin{equation}
        \begin{aligned}
        x^{t}_{i} &= \textrm{MHA}(x_i, \bar{H}, \bar{H}), \\
        \bar{x}_{i} &= x^{t}_{i} + x_i,
        \end{aligned}
    \end{equation}
where $\textrm{MHA}$ denotes the multi-head attention mechanism and $\bar{H}$ denotes the embeddings of all nodes in the history semantic graph.

    \item \textbf{Utterance-level Aggregation}: 
    Sometimes the token itself may not contain semantic information, e.g., stop words. We further propose to incorporate history information at the utterance-level for these tokens:
    \begin{equation}
        \begin{aligned}
        x^{u}_{i} &= \textrm{MHA}(x_{\textrm{[CLS]}}, \bar{H}, \bar{H}), \\
        \bar{x}_{i} &= x^{u}_{i} + x_i,
        \end{aligned}
    \end{equation}
where $x_{\textrm{[CLS]}}$ denotes the representation of the \texttt{[CLS]} token.
\end{itemize}

Then, history-semantic-aware token embeddings are forwarded as input to the encoder of Transformer~\cite{vaswani2017attention} for deep interaction:
\begin{equation}
    \begin{aligned}
        h^{(enc)} &= \textrm{Encoder}(\bar{X}; \theta^{(enc)}),
    \end{aligned}
\end{equation}
where $\theta^{(enc)}$ are encoder trainable parameters.

\subsection{Grammar-Guided Decoder}
After encoding all the semantic information into the hidden state $h^{(enc)}$, we utilize stacked masked attention mechanism~\cite{vaswani2017attention} to generate sequence-formatted logical forms. 
Specifically, in each decoding step, our model predicts a token from a small decoding vocabulary $V^{(dec)} = \{ start, end, e, p, tp,..., find\}$, where all the actions from the \cref{table:grammar} are included. On top of the decoder, we employ a linear layer alongside a softmax to calculate each token's probability distribution in the vocabulary. The detailed computation is defined as follows:
\begin{equation}
    \begin{aligned}
        h^{(dec)} &= \textrm{Decoder}(h^{(enc)}; \theta^{(dec)}), \\
        p_t^{(dec)} &= \textrm{Softmax}(W^{(dec)} h_t^{(dec)}),
    \end{aligned}
\end{equation}
where $h_t^{(dec)}$ is the hidden state at time step $t$, $\theta^{(dec)}, W^{(dec)}$ are decoder trainable parameters, $p_t^{(dec)} \in \mathbb{R}^{|V^{(dec)}|}$ is the probability distribution over the decoding vocabulary at time step $t$.

\subsection{Entity Recognition Module}
Entity recognition module aims to fill the entity slot in the predicted logical forms, which consists of entity detection module and entity linking module.

\paragraph{Entity Detection.}
The goal of entity detection is to identify mentions of entities in the input. Previous studies~\cite{shen2019multi} have shown that multiple entities of different types in a large KB may share the same entity text, which is a common phenomenon called \texttt{Named Entity Ambiguity}. To address this issue and inspired by \cite{kacupaj2021conversational}, we adopt a type-aware entity detection approach using BIO sequence tagging. Specifically, the entity detection vocabulary is defined as $V^{(ed)} = \{O, \{B, I\} \times \{TP_i\}_{i = 1}^{N^{(tp)}} \}$, where $TP_i$ denotes the $i$-th entity type label, $N^{(tp)}$ stands for the number of distinct entity types in the knowledge graph and $|V^{(ed)}| = 2 \times N^{(tp)} + 1$. We leverage LSTM~\cite{hochreiter1997long} to perform the sequence tagging task:
\begin{equation}
    \begin{aligned}
        h^{(ed)} &= \textrm{LeakyReLU}(\textrm{LSTM}(h^{(enc)}; \theta^{(l)})),\\
        p_t^{(ed)} &= \textrm{Softmax}(W^{(ed)}h_t^{(ed)}),
    \end{aligned}
\end{equation}
where $h^{(enc)}$ is the encoder hidden state, $\theta^{(l)}$ are the LSTM trainable parameters, $h_t^{(ed)}$ is the LSTM hidden state at time step $t$, and $p_t^{(ed)}$ is the probability  distribution over $V^{(ed)}$ at time step $t$.

\paragraph{Entity Linking.}
Once we detect the entities in the input utterance, we perform entity linking to link the entities to the entity slots in the predicted logical form. Specifically, we define the entity linking vocabulary as $V^{(el)} = \{0,1,...,M\}$ where $0$ means that the entity does not link to any entity slot in the predicted logical form and $M$ denotes the total number of indices based on the maximum number of entities from all logical forms. The probability distribution is defined as follows:
\begin{equation}
    \begin{aligned}
        h^{(el)} &= \textrm{LeakyReLU}(W^{(el_1)}[h^{(enc)}; h^{(ed)}]),\\
        p_t^{(el)} &= \textrm{Softmax}(W^{(el_2)}h_t^{(el)}),
    \end{aligned}
\end{equation}
where $W^{(el_1)}, W^{(el_2)}$ are trainable parameters, $h_t^{(el)}$ is the hidden state at time step $t$, and $p_t^{(el)}$ is the probability distribution over the tag indices $V^{(el)}$ at time step $t$.

\subsection{Concept-aware Attention Module}
In the Concept-aware Attention Module, we first model the complex interaction between entity types and predicates, then we predict the entity types and predicates for the logical form.

To begin with, we first develop an entity-to-concept converter to replace the entities in each factual triple of Wikidata KG with corresponding concepts (i.e., entity types). Take an instance in \cref{image:graph} as example, the factual triple (Joe Biden, IsPresidentOf, USA) can be transformed to two concept-level tuples (Person, IsPresidentOf),  and (IsPresidentOf, Country) in the concept graph. Then, we initialize node embeddings using their texts with BERT and apply Graph Attention Networks (GAT)~\cite{velivckovic2017graph} to project the KG information into the embedding space.

Finally, we model the task of predicting the correct entity type or predicate of the logical form as a classification task. For each time step of decoding, we directly calculate the probability distribution at time step $t$ as:
\begin{equation}
    \begin{aligned}
        h_t^{(c)} &= \textrm{LeakyReLU}(W^{(c)}[h^{(enc)}_{\textrm{[CLS]}}; h_t^{(dec)}]),\\
        p_t^{(c)} &= \textrm{Softmax}(h^{(g)T}h_t^{(c)}),
    \end{aligned}
\end{equation}
where $h^{(g)}$ is the updated entity type and predicate embedding and $p_t^{(c)}$ is the probability distribution over them at time step $t$.

\subsection{Training}
The framework consists of four trainable modules: Entity Detection Module, Entity Linking Module, Grammar-guided Decoder and Concept-aware Attention Module. Each module consists of a loss function that can be used to optimize the parameters in itself. We use the weighted average of all the losses as our loss function:
\begin{equation}
    L = \lambda_1 L^{ed} + \lambda_2 L^{el} + \lambda_3 L^{dec} + \lambda_4 L^{c},
    \label{eq:loss}
\end{equation}
where $\lambda_1, \lambda_2, \lambda_3, \lambda_4$ are the weights that decide the importance of each component. The detailed loss calculation method is in \cref{sec:loss}. The multi-task setting enables modules to share supervision signals, which benefits the model performance.






\section{Experiments}

\subsection{Experimental Setup}
\paragraph{Dataset.}
We conduct experiments on CSQA (Complex Sequential Question Answering) dataset~\footnote{\href{https://amritasaha1812.github.io/CSQA/}{https://amritasaha1812.github.io/CSQA}} \cite{saha2018complex}. CSQA was built based on the Wikidata knowledge graph, which consists of 21.1M triples with over 12.8M entities, 3,054 entity types and 567 predicates. CSQA dataset is the largest dataset for conversational KBQA and consists of around 200K dialogues where training set, validation set and testing set contain 153K, 16K and 28K dialogues, respectively. Questions in the dataset are classified as different types, e.g., simple questions, logical reasoning and so on.
\paragraph{Metrics.}
To evaluate HSGE, We use the same metrics as employed by the authors of the CSQA dataset as well as the previous baselines. \textbf{F1 score} is used to evaluate the question whose answer is comprised of entities, while \textbf{Accuracy} is used to measure the question whose answer is a number or a boolean number.
Following \cite{marion2021structured}, we don’t report results for ``Clarification'' question type, as this question type can be accurately modeled with a simple classification task.

\paragraph{Baselines.}
We compare HSGE with the latest five baselines that include D2A~\cite{guo2018dialog}, S2A+MAML~\cite{guo2019coupling}, MaSP~\cite{shen2019multi}, OAT~\cite{marion2021structured} and LASAGNE~\cite{kacupaj2021conversational}.

\subsection{Overall Performance}
\begin{table*}[t]
\centering
\scalebox{0.97}{
\setlength{\tabcolsep}{2mm}{
\begin{tabular}{l|c|cccccc}
\toprule
\textbf{Methods}     &      & D2A   & S2A-MAML  & MaSP    & OAT  & LASAGNE &  HSGE   \\ \hline
\textbf{Question Type} & \textbf{\#Example} & \multicolumn{6}{c}{F1 Score}         \\ \hline
Comparative & 15K     & 44.20  & 48.13  & 68.90  &  \textbf{70.76}  & 69.77    &  69.70       \\
Logical & 22K     & 43.62  & 44.34  & 69.04  &  81.57 &  89.83       &  \textbf{91.24}  \\
Quantitative    & 9K       & 50.25  & 50.30  & 73.75  &  74.83 &  86.67       &  \textbf{87.37}   \\
Simple (Coreferenced) & 55K  & 69.83  & 71.18  & 76.47   & \textbf{79.23}&  79.06       &  78.73  \\
Simple (Direct) & 82K  &  91.41   & \textbf{92.66}  & 85.18  & 82.69 &  87.95      &89.38       \\
Simple (Ellipsis)     & 10K   & 81.98  & 82.21  &  83.73  & \textbf{84.44} & 80.09     &   80.53    \\ \hline
\textbf{Question Type}                  & \textbf{\#Example} & \multicolumn{6}{c}{Accuracy}          \\ \hline
Verification (Boolean)& 27K  & 45.05  & 50.16  & 60.63  & 66.39  &  78.86      &   \textbf{82.17}         \\
Quantitative (Count) & 24K   & 40.94  & 46.43 & 43.39  &  71.79  & 55.18     &   \textbf{72.88 } \\
Comparative (Count)  & 15K    & 17.78  & 18.91 & 22.26  & 36.00 &  53.34     &  \textbf{53.74}       \\ \hline \hline
\textbf{Overall}  & 260K   & 64.47  & 66.54 & 70.56  & 75.57 & 78.82   &  \textbf{81.38}\bm{$^{\ast\dagger\S}$}       \\
\bottomrule
\end{tabular}
 }}
\caption{HSGE’s performance comparison on the CSQA dataset. HSGE achieves new state-of-the-art on the overall performance averaged on all question types. We use the paired t-test with $p\leq0.01$. The superscripts refer to significant improvements compared to LASAGNE($^\ast$), OAT($^\dagger$) and MaSP($^\S$).}
\label{table:overall}
\end{table*}
\cref{table:overall} summarizes the results comparing the HSGE framework against the previous baselines. From the result, we have three observations:

(1) The D2A and S2A-MAML models exhibit superior performance on the \emph{Simple Question (Direct)} question type. This can likely be attributed to their ability to memorize context information previously mentioned in the conversation. However, these models fail to model the complex interaction between entities, resulting in inferior performance on other question types.

(2) OAT achieves superior performance on three question types, which might be attributed to its incorporation of additional KG information. However, its performance is not consistent across all question types, leading to a low overall performance averaged on all question types.

(3) Our method HSGE achieves the new SOTA on the overall performance averaged on all question types.
There are two possible reasons for the improvement. First, the incorporation of HSG allows the modeling of longer dependencies within the context, enabling the model to handle situations where the user asks about entities that were previously mentioned. Second, by utilizing graph neural network to facilitate information flow in HSG, the interaction among previously appeared entities, entity types and predicates are better captured, which endows our model with stronger reasoning ability.

\subsection{Ablation Study}
In this section, we first conduct experiments to verify the effectiveness of each model component. Then, we investigate the effects of different model choices inside the Context-aware Encoder. Finally, we compare our HSGE with the most widely used concatenation method.

\paragraph{Effect of HSG and TIM.}
\begin{table}[t]
\centering
\scalebox{0.85}{
\begin{tabular}{l|ccc}
\hline
Methods           & Ours  & w/o HSG  & w/o TIM \\ \hline
Question Type     & \multicolumn{3}{c}{F1 Score} \\ \hline
Comparative       &   \textbf{69.70}  &  69.47  &  69.55  \\
Logical           &   \textbf{91.24}  &  87.99  &  89.99  \\
Quantitative      &   \textbf{87.37}  &  86.63  &  86.71  \\
Simple (Coref)    &   \textbf{78.73}  &  77.78  &  78.17  \\
Simple (Direct)   &   \textbf{89.38}  &  88.64  &  88.97  \\
Simple (Ellipsis) &   \textbf{80.53}  &  78.60  &  79.95  \\\hline
Question Type     &  \multicolumn{3}{c}{Accuracy}      \\ \hline
Verification      &   \textbf{82.17}  &  79.70  &  78.05  \\
Quantitative (Count) &   \textbf{72.88}  &  69.00  &  71.29  \\
Comparative (Count) &   \textbf{53.74}  &  52.70  &  53.14  \\\hline \hline
Overall     &   \textbf{81.38}\bm{$^{\ast\dagger}$}  & 79.87  &   80.36 \\\hline
\end{tabular}
}
\caption{Ablation Study. We use the paired t-test with $p\leq0.01$. The superscripts refer to significant improvements compared to w/o HSG($^\ast$) and w/o TIM($^\dagger$).}
\label{table:ablation}
\end{table}
To show the effectiveness of each component, we create two ablations by directly removing history semantic graph (HSG) and temporal information modeling (TIM), respectively. As shown in \cref{table:ablation}, HSGE outperforms all the ablations across all question types, which verifies the importance of each model component.

It is worth mentioning that after removing HSG, the performance of our method on some question types that require reasoning (i.e., \emph{Logical Reasoning, Quantitative Reasoning (Count)}) drops significantly. We think that the reason might be the utilization of graph neural network on HSG empowers the model with great reasoning ability, which further benefits model performance.

\paragraph{Comparison of Internal Model Choice.}

\begin{figure}[htbp]
\centering
\includegraphics[width=7.5cm]{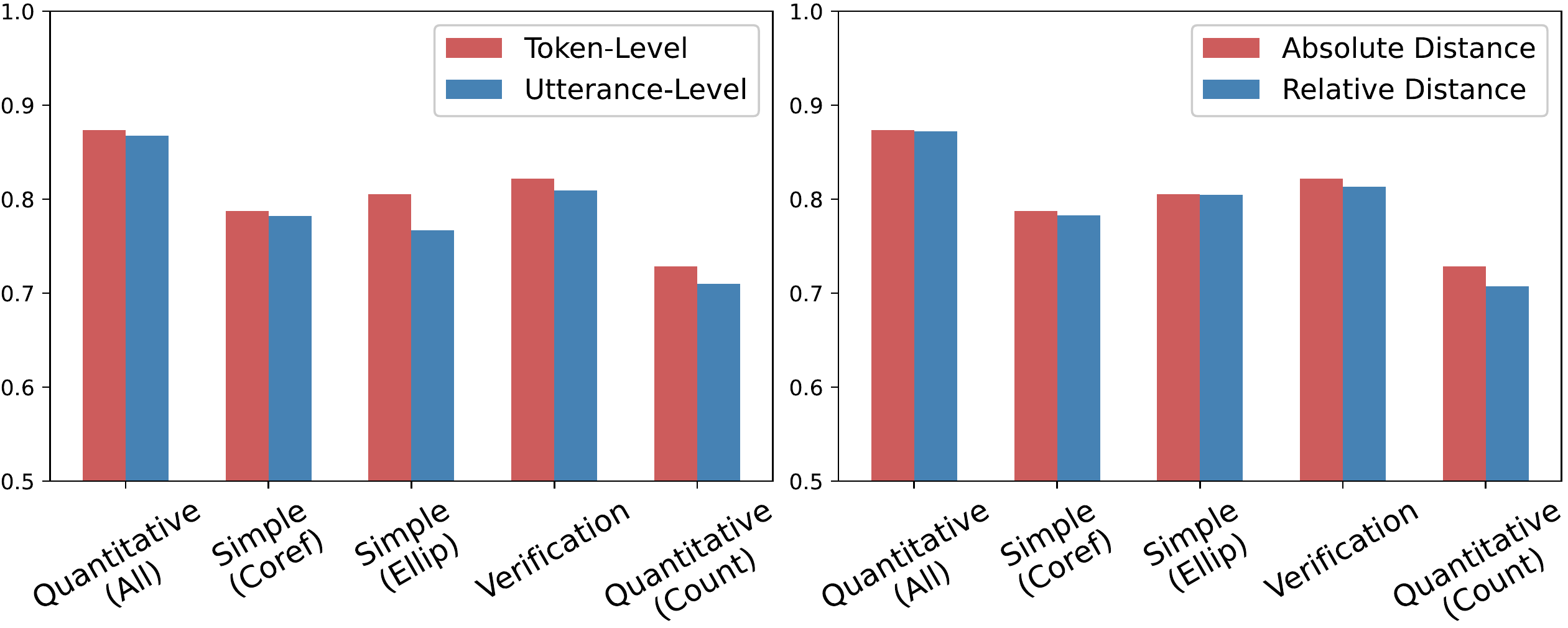}
\caption{The comparison between token/utterance-level aggregation and between absolute/relative distance on five selected question types.}
\label{image:comparison}
\end{figure}
In context-aware encoder, we design two distance calculation methods (i.e., absolute distance and relative distance) for temporal information modeling, as well as two information aggregation granularities (i.e., token-level and utterance-level aggregation) for semantic information aggregation. To study their effects, we conduct experiments by fixing one setting while changing the other. And the comparison result is shown in \cref{image:comparison}. 

From the results,  it is obvious that we can get the following conclusions:
(1) Token-level aggregation method performs better than utterance-level aggregation method. This is because the token-level aggregation allows the model to incorporate context information at a finer granularity and the information unrelated to the target token can be removed.
(2) Absolute distance method performs better than relative distance method. The reason may be that although both distance calculation methods can provide temporal information, absolute distance is more informative since the model can derive relative distance using absolute distance while the opposite is not true.

\paragraph{Comparison with Concatenation Method.}
\begin{figure}[htbp]
\centering
\includegraphics[width=7.5cm]{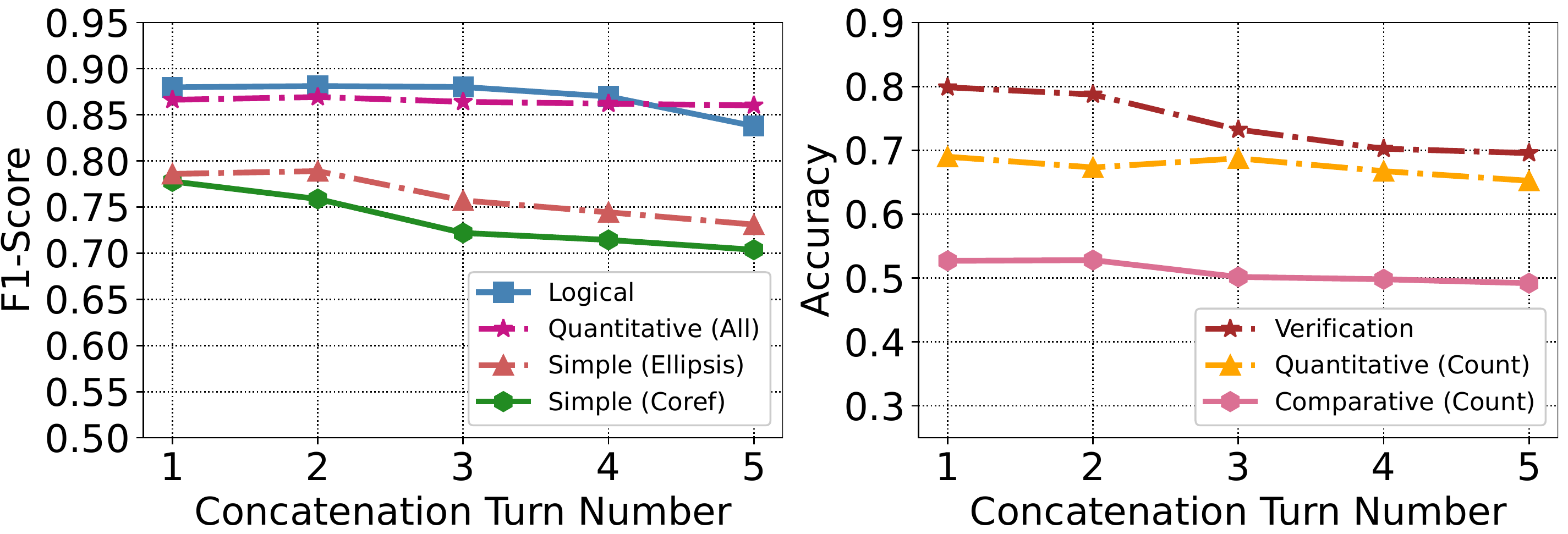}
\caption{The performance of the concatenation method on seven representative question types with regard to the concatenation turn number.}
\label{image:turns}
\end{figure}
One of the most widely used methods for context modeling is to directly concatenate history conversations~\cite{liu2020far}. 
To analyze its effectiveness, we remove HSG and observe the performance of seven representative question types using the concatenation of history conversations as input, which is shown in \cref{image:turns}.

As we can see, at the initial stages of concatenation turn number increase, the performances on some question types increase a little while remaining unchanged or even decreasing on others, leading to an almost unchanged overall performance.
It is reasonable because history turns contain useful semantic information, which leads to performance gain. 
However, as more conversation turns are introduced into the model, more noisy tokens will also be introduced into the model, which leads to performance degradation. Besides, the introduction of more context tokens will also lead to an increase in computational cost with the $O(N^2)$ complexity.

It is worth noting that the best setting of concatenation method still performs worse than HSGE. It is mainly because we use attention mechanism to dynamically select the most related entities from the HSG, which achieves effective history modeling while avoiding introducing noisy information. And as we only extract entities and predicates from history conversations, the size of the graph is relatively small and the increase in computational cost as the conversation progresses is marginal.

\subsection{Subtask Analysis}
\begin{table}[t]
\centering
\scalebox{0.9}{
\begin{tabular}{c|cc}
\hline
Task                                     & LASAGNE & HSGE    \\ \hline
Entity Detection                         & 86.75\% & \textbf{89.75\%} \\
Entity Linking                           & 97.49\% & \textbf{98.19\%} \\
Logical Form Generation                  & \textbf{98.61\%} & 92.76\% \\
Type\&Predicate Prediction     & 92.28\% & \textbf{93.11\%} \\ \hline
\end{tabular}}
\caption{Comparison of subtask accuracy in LASAGNE and HSGE.}
\label{table:task}
\end{table}
The task of conversational KBQA involves multiple subtasks, each of which can directly impact the final model accuracy. To gain a deeper understanding of HSGE, we compare its performance of each subtask with the current SOTA model LASAGNE in \cref{table:task}.
We can observe that most of the sub-task's performance in HSGE is better than that of LASAGNE and mostly achieves accuracy above 90\%. Amongst them, the improvement in Entity Detection is the largest. We think the main reason is that the token-level aggregation mechanism endows each token with richer semantic information.

\subsection{Error Analysis}
In this section, we randomly sample 200 incorrect predictions and analyze their error causes:

\paragraph{Entity Ambiguity.}
Entity ambiguity refers to the situation where there exist multiple entities with the same text and type in the Wikidata knowledge graph. For example, we cannot distinguish multiple people called ``Mary Johnson'' because we have no more information other than entity text and entity type. We believe that incorporating other contextual information such as entity descriptions  may help solve this problem~\cite{mulang2020encoding}.

\paragraph{Spurious Logical Form.}
We follow \cite{shen2019multi,kacupaj2021conversational} and produce golden logical forms by leveraging BFS to search valid logical forms for questions in training data. This can sometimes lead to wrong golden actions such as two actions with different semantic information but accidentally sharing the same execution result. This may misguide our model during training.

\section{Conclusion}
In this paper, we propose a novel Conversational KBQA method HSGE, which achieves effective history modeling with minimal computational cost. We design a context-aware encoder that introduces temporal embedding to address user's conversation focus shift phenomenon and aggregate context information at both token-level and utterance-level. Our proposed HSGE outperforms existing baselines averaged on all question types on the widely used CSQA dataset.




\bibliographystyle{acl_natbib}
\bibliography{HSGE}

\begin{thebibliography}{36}
\expandafter\ifx\csname natexlab\endcsname\relax\def\natexlab#1{#1}\fi

\bibitem[{Androutsopoulos et~al.(1995)Androutsopoulos, Ritchie, and
  Thanisch}]{androutsopoulos1995natural}
Ion Androutsopoulos, Graeme~D Ritchie, and Peter Thanisch. 1995.
\newblock Natural language interfaces to databases--an introduction.
\newblock \emph{Natural language engineering}, 1(1):29--81.

\bibitem[{Artzi and Zettlemoyer(2013)}]{artzi2013weakly}
Yoav Artzi and Luke Zettlemoyer. 2013.
\newblock Weakly supervised learning of semantic parsers for mapping
  instructions to actions.
\newblock \emph{Transactions of the Association for Computational Linguistics},
  1:49--62.

\bibitem[{Auer et~al.(2007)Auer, Bizer, Kobilarov, Lehmann, Cyganiak, and
  Ives}]{auer2007dbpedia}
S{\"o}ren Auer, Christian Bizer, Georgi Kobilarov, Jens Lehmann, Richard
  Cyganiak, and Zachary Ives. 2007.
\newblock Dbpedia: A nucleus for a web of open data.
\newblock In \emph{The semantic web}, pages 722--735. Springer.

\bibitem[{Bollacker et~al.(2008)Bollacker, Evans, Paritosh, Sturge, and
  Taylor}]{bollacker2008freebase}
Kurt Bollacker, Colin Evans, Praveen Paritosh, Tim Sturge, and Jamie Taylor.
  2008.
\newblock Freebase: a collaboratively created graph database for structuring
  human knowledge.
\newblock In \emph{Proceedings of the 2008 ACM SIGMOD international conference
  on Management of data}, pages 1247--1250.

\bibitem[{Chen et~al.(2019)Chen, Wu, and Zaki}]{chen2019graphflow}
Yu~Chen, Lingfei Wu, and Mohammed~J Zaki. 2019.
\newblock Graphflow: Exploiting conversation flow with graph neural networks
  for conversational machine comprehension.
\newblock \emph{arXiv preprint arXiv:1908.00059}.

\bibitem[{Das et~al.(2021)Das, Zaheer, Thai, Godbole, Perez, Lee, Tan,
  Polymenakos, and McCallum}]{das2021case}
Rajarshi Das, Manzil Zaheer, Dung Thai, Ameya Godbole, Ethan Perez, Jay-Yoon
  Lee, Lizhen Tan, Lazaros Polymenakos, and Andrew McCallum. 2021.
\newblock Case-based reasoning for natural language queries over knowledge
  bases.
\newblock \emph{arXiv preprint arXiv:2104.08762}.

\bibitem[{Devlin et~al.(2018)Devlin, Chang, Lee, and
  Toutanova}]{devlin2018bert}
Jacob Devlin, Ming-Wei Chang, Kenton Lee, and Kristina Toutanova. 2018.
\newblock Bert: Pre-training of deep bidirectional transformers for language
  understanding.
\newblock \emph{arXiv preprint arXiv:1810.04805}.

\bibitem[{Dong and Lapata(2016)}]{dong2016language}
Li~Dong and Mirella Lapata. 2016.
\newblock Language to logical form with neural attention.
\newblock \emph{arXiv preprint arXiv:1601.01280}.

\bibitem[{Guo et~al.(2018)Guo, Tang, Duan, Zhou, and Yin}]{guo2018dialog}
Daya Guo, Duyu Tang, Nan Duan, Ming Zhou, and Jian Yin. 2018.
\newblock Dialog-to-action: Conversational question answering over a
  large-scale knowledge base.
\newblock \emph{Advances in Neural Information Processing Systems}, 31.

\bibitem[{Guo et~al.(2019)Guo, Tang, Duan, Zhou, and Yin}]{guo2019coupling}
Daya Guo, Duyu Tang, Nan Duan, Ming Zhou, and Jian Yin. 2019.
\newblock Coupling retrieval and meta-learning for context-dependent semantic
  parsing.
\newblock \emph{arXiv preprint arXiv:1906.07108}.

\bibitem[{Hochreiter and Schmidhuber(1997)}]{hochreiter1997long}
Sepp Hochreiter and J{\"u}rgen Schmidhuber. 1997.
\newblock Long short-term memory.
\newblock \emph{Neural computation}, 9(8):1735--1780.

\bibitem[{Hui et~al.(2021)Hui, Geng, Ren, Li, Li, Sun, Huang, Si, Zhu, and
  Zhu}]{hui2021dynamic}
Binyuan Hui, Ruiying Geng, Qiyu Ren, Binhua Li, Yongbin Li, Jian Sun, Fei
  Huang, Luo Si, Pengfei Zhu, and Xiaodan Zhu. 2021.
\newblock Dynamic hybrid relation network for cross-domain context-dependent
  semantic parsing.
\newblock \emph{arXiv preprint arXiv:2101.01686}.

\bibitem[{Iyyer et~al.(2017)Iyyer, Yih, and Chang}]{iyyer-etal-2017-search}
Mohit Iyyer, Wen-tau Yih, and Ming-Wei Chang. 2017.
\newblock Search-based neural structured learning for sequential question
  answering.
\newblock In \emph{ACL}.

\bibitem[{Kacupaj et~al.(2021)Kacupaj, Plepi, Singh, Thakkar, Lehmann, and
  Maleshkova}]{kacupaj2021conversational}
Endri Kacupaj, Joan Plepi, Kuldeep Singh, Harsh Thakkar, Jens Lehmann, and
  Maria Maleshkova. 2021.
\newblock Conversational question answering over knowledge graphs with
  transformer and graph attention networks.
\newblock \emph{arXiv preprint arXiv:2104.01569}.

\bibitem[{Lan and Jiang(2021)}]{lan2021modeling}
Yunshi Lan and Jing Jiang. 2021.
\newblock Modeling transitions of focal entities for conversational knowledge
  base question answering.
\newblock In \emph{Proceedings of the 59th Annual Meeting of the Association
  for Computational Linguistics and the 11th International Joint Conference on
  Natural Language Processing (Volume 1: Long Papers)}, pages 3288--3297.

\bibitem[{Liang et~al.(2016)Liang, Berant, Le, Forbus, and
  Lao}]{liang2016neural}
Chen Liang, Jonathan Berant, Quoc Le, Kenneth~D Forbus, and Ni~Lao. 2016.
\newblock Neural symbolic machines: Learning semantic parsers on freebase with
  weak supervision.
\newblock \emph{arXiv preprint arXiv:1611.00020}.

\bibitem[{Liu et~al.(2020)Liu, Chen, Guo, Lou, Zhou, and Zhang}]{liu2020far}
Qian Liu, Bei Chen, Jiaqi Guo, Jian-Guang Lou, Bin Zhou, and Dongmei Zhang.
  2020.
\newblock How far are we from effective context modeling? an exploratory study
  on semantic parsing in context.
\newblock \emph{arXiv preprint arXiv:2002.00652}.

\bibitem[{Long et~al.(2016)Long, Pasupat, and Liang}]{long2016simpler}
Reginald Long, Panupong Pasupat, and Percy Liang. 2016.
\newblock Simpler context-dependent logical forms via model projections.
\newblock \emph{arXiv preprint arXiv:1606.05378}.

\bibitem[{Marion et~al.(2021)Marion, Nowak, and
  Piccinno}]{marion2021structured}
Pierre Marion, Pawe{\l}~Krzysztof Nowak, and Francesco Piccinno. 2021.
\newblock Structured context and high-coverage grammar for conversational
  question answering over knowledge graphs.
\newblock \emph{arXiv preprint arXiv:2109.00269}.

\bibitem[{Mulang et~al.(2020)Mulang, Singh, Vyas, Shekarpour, Vidal, Lehmann,
  and Auer}]{mulang2020encoding}
Isaiah~Onando Mulang, Kuldeep Singh, Akhilesh Vyas, Saeedeh Shekarpour,
  Maria-Esther Vidal, Jens Lehmann, and Soren Auer. 2020.
\newblock Encoding knowledge graph entity aliases in attentive neural network
  for wikidata entity linking.
\newblock In \emph{International Conference on Web Information Systems
  Engineering}, pages 328--342. Springer.

\bibitem[{Plepi et~al.(2021)Plepi, Kacupaj, Singh, Thakkar, and
  Lehmann}]{plepi2021context}
Joan Plepi, Endri Kacupaj, Kuldeep Singh, Harsh Thakkar, and Jens Lehmann.
  2021.
\newblock Context transformer with stacked pointer networks for conversational
  question answering over knowledge graphs.
\newblock In \emph{European Semantic Web Conference}, pages 356--371. Springer.

\bibitem[{Qu et~al.(2019)Qu, Yang, Qiu, Zhang, Chen, Croft, and
  Iyyer}]{qu2019attentive}
Chen Qu, Liu Yang, Minghui Qiu, Yongfeng Zhang, Cen Chen, W~Bruce Croft, and
  Mohit Iyyer. 2019.
\newblock Attentive history selection for conversational question answering.
\newblock In \emph{Proceedings of the 28th ACM International Conference on
  Information and Knowledge Management}, pages 1391--1400.

\bibitem[{Saha et~al.(2018)Saha, Pahuja, Khapra, Sankaranarayanan, and
  Chandar}]{saha2018complex}
Amrita Saha, Vardaan Pahuja, Mitesh Khapra, Karthik Sankaranarayanan, and
  Sarath Chandar. 2018.
\newblock Complex sequential question answering: Towards learning to converse
  over linked question answer pairs with a knowledge graph.
\newblock In \emph{Proceedings of the AAAI Conference on Artificial
  Intelligence}, volume~32.

\bibitem[{Shen et~al.(2019)Shen, Geng, Qin, Guo, Tang, Duan, Long, and
  Jiang}]{shen2019multi}
Tao Shen, Xiubo Geng, Tao Qin, Daya Guo, Duyu Tang, Nan Duan, Guodong Long, and
  Daxin Jiang. 2019.
\newblock Multi-task learning for conversational question answering over a
  large-scale knowledge base.
\newblock \emph{arXiv preprint arXiv:1910.05069}.

\bibitem[{Shi et~al.(2020)Shi, Huang, Feng, Zhong, Wang, and
  Sun}]{shi2020masked}
Yunsheng Shi, Zhengjie Huang, Shikun Feng, Hui Zhong, Wenjin Wang, and Yu~Sun.
  2020.
\newblock Masked label prediction: Unified message passing model for
  semi-supervised classification.
\newblock \emph{arXiv preprint arXiv:2009.03509}.

\bibitem[{Suhr et~al.(2018)Suhr, Iyer, and Artzi}]{suhr-etal-2018-learning}
Alane Suhr, Srinivasan Iyer, and Yoav Artzi. 2018.
\newblock Learning to map context-dependent sentences to executable formal
  queries.
\newblock In \emph{NAACL}.

\bibitem[{Vaswani et~al.(2017)Vaswani, Shazeer, Parmar, Uszkoreit, Jones,
  Gomez, Kaiser, and Polosukhin}]{vaswani2017attention}
Ashish Vaswani, Noam Shazeer, Niki Parmar, Jakob Uszkoreit, Llion Jones,
  Aidan~N Gomez, {\L}ukasz Kaiser, and Illia Polosukhin. 2017.
\newblock Attention is all you need.
\newblock \emph{Advances in neural information processing systems}, 30.

\bibitem[{Veli{\v{c}}kovi{\'c} et~al.(2017)Veli{\v{c}}kovi{\'c}, Cucurull,
  Casanova, Romero, Lio, and Bengio}]{velivckovic2017graph}
Petar Veli{\v{c}}kovi{\'c}, Guillem Cucurull, Arantxa Casanova, Adriana Romero,
  Pietro Lio, and Yoshua Bengio. 2017.
\newblock Graph attention networks.
\newblock \emph{arXiv preprint arXiv:1710.10903}.

\bibitem[{Wang et~al.(2020)Wang, Zhao, Han, Cheng, Yang, Ao, and
  Li}]{wang2020modelling}
Xu~Wang, Shuai Zhao, Jiale Han, Bo~Cheng, Hao Yang, Jianchang Ao, and Zhenzi
  Li. 2020.
\newblock Modelling long-distance node relations for kbqa with global dynamic
  graph.
\newblock In \emph{Proceedings of the 28th International Conference on
  Computational Linguistics}, pages 2572--2582.

\bibitem[{Wang et~al.(2022)Wang, Jin et~al.}]{wang2022new}
Yu~Wang, Hongxia Jin, et~al. 2022.
\newblock A new concept of knowledge based question answering (kbqa) system for
  multi-hop reasoning.
\newblock In \emph{Proceedings of the 2022 Conference of the North American
  Chapter of the Association for Computational Linguistics: Human Language
  Technologies}, pages 4007--4017.

\bibitem[{Wu et~al.(2016)Wu, Schuster, Chen, Le, Norouzi, Macherey, Krikun,
  Cao, Gao, Macherey et~al.}]{wu2016google}
Yonghui Wu, Mike Schuster, Zhifeng Chen, Quoc~V Le, Mohammad Norouzi, Wolfgang
  Macherey, Maxim Krikun, Yuan Cao, Qin Gao, Klaus Macherey, et~al. 2016.
\newblock Google's neural machine translation system: Bridging the gap between
  human and machine translation.
\newblock \emph{arXiv preprint arXiv:1609.08144}.

\bibitem[{Yadati et~al.(2021)Yadati, Dayanidhi, Vaishnavi, Indira, and
  Srinidhi}]{yadati2021knowledge}
Naganand Yadati, RS~Dayanidhi, S~Vaishnavi, KM~Indira, and G~Srinidhi. 2021.
\newblock Knowledge base question answering through recursive hypergraphs.
\newblock In \emph{Proceedings of the 16th Conference of the European Chapter
  of the Association for Computational Linguistics: Main Volume}, pages
  448--454.

\bibitem[{Yan et~al.(2021)Yan, Li, Wang, Zhang, Daoguang, Zhang, Wu, and
  Xu}]{yan2021large}
Yuanmeng Yan, Rumei Li, Sirui Wang, Hongzhi Zhang, Zan Daoguang, Fuzheng Zhang,
  Wei Wu, and Weiran Xu. 2021.
\newblock Large-scale relation learning for question answering over knowledge
  bases with pre-trained language models.
\newblock In \emph{Proceedings of the 2021 Conference on Empirical Methods in
  Natural Language Processing}, pages 3653--3660.

\bibitem[{Ye et~al.(2021)Ye, Yavuz, Hashimoto, Zhou, and Xiong}]{ye2021rng}
Xi~Ye, Semih Yavuz, Kazuma Hashimoto, Yingbo Zhou, and Caiming Xiong. 2021.
\newblock Rng-kbqa: Generation augmented iterative ranking for knowledge base
  question answering.
\newblock \emph{arXiv preprint arXiv:2109.08678}.

\bibitem[{Yu et~al.(2019)Yu, Zhang, Er, Li, Xue, Pang, Lin, Tan, Shi, Li,
  Jiang, Yasunaga, Shim, Chen, Fabbri, Li, Chen, Zhang, Dixit, Zhang, Xiong,
  Socher, Lasecki, and Radev}]{yu-etal-2019-cosql}
Tao Yu, Rui Zhang, Heyang Er, Suyi Li, Eric Xue, Bo~Pang, Xi~Victoria Lin,
  Yi~Chern Tan, Tianze Shi, Zihan Li, Youxuan Jiang, Michihiro Yasunaga,
  Sungrok Shim, Tao Chen, Alexander Fabbri, Zifan Li, Luyao Chen, Yuwen Zhang,
  Shreya Dixit, Vincent Zhang, Caiming Xiong, Richard Socher, Walter Lasecki,
  and Dragomir Radev. 2019.
\newblock {C}o{SQL}: A conversational text-to-{SQL} challenge towards
  cross-domain natural language interfaces to databases.
\newblock In \emph{EMNLP-IJCNLP}.

\bibitem[{Zettlemoyer and Collins(2009)}]{zettlemoyer2009learning}
Luke~S Zettlemoyer and Michael Collins. 2009.
\newblock Learning context-dependent mappings from sentences to logical form.

\end{thebibliography}

\newpage
\appendix
\section{Grammar}
\begin{table*}[ht]
\centering
\scalebox{0.77}{
\setlength{\tabcolsep}{2mm}{
\begin{tabular}{lp{12cm}}
\toprule 
\textbf{Action}                             & \textbf{Description}\\ \hline
$set \rightarrow find(e, p)$                              & set of subjects part of the triples with object $e$ and predicate $p$\\
$set \rightarrow find\_reverse(e, p)$                      & set of objects part of the triples with subject $e$ and predicate $p$\\
$set \rightarrow filter\_type(set, tp)$                    & filter the given set of entities based on the given type\\
$set \rightarrow filter\_multi\_types(set_1, set_2)$     & filter the given set of entities based on the given set of types\\
$dict \rightarrow find\_tuple\_counts(p, tp_1, tp_)$         & extracts a dictionary, where keys are entities of $type_1$ and values are the number of objects of $type_2$ related with $p$  \\
$dict \rightarrow find\_reverse\_tuple\_counts(p, tp_1, tp_2)$ & extracts a dictionary, where keys are entities of $type_1$ and values are the number of subjects of $type_2$ related with $p$ \\
$set \rightarrow greater(dict, num)$                      & set of those entities that have lesser count than $num$\\
$set \rightarrow lesser(dict, num)$                       & set of those entities that have greater count than $num$\\
$set \rightarrow equal(dict, num)$                        & set of those entities that have equal count with $num$\\
$set \rightarrow approx(dict, num)$                       & set of those entities that have approximately same count with $num$\\
$set \rightarrow atmost(dict, num)$                       & set of those entities that have at most same count with $num$\\
$set \rightarrow atleast(dict, num)$                      & set of those entities that have at least same count with $num$\\
$set \rightarrow argmin(dict)$                            & set of those entities that have the most count\\
$set \rightarrow argmax(dict)$                            & set of those entities that have the least count\\
$boolean \rightarrow is\_in(entity, set)$                  & check if the entity is part of the $set$\\
$number \rightarrow count(set)$                           & count the number of elements in the $set$\\
$set \rightarrow union(set_1, set_2)   $                    & union of $set_1$ and $set_2$\\
$set \rightarrow intersection(set_1, set_2)  $              & intersection of $set_1$ and $set_2$\\
$set \rightarrow difference(set_1, set_2)$                & difference of $set_1$ and $set_2$ \\       \bottomrule
\end{tabular}
}}
\caption{The grammar we use in this work for generating logical forms.}
\label{table:grammar}
\end{table*}

The grammar we use in this work is defined in \cref{table:grammar}. Please note that each single action can only model relatively simple semantics. High-level semantics of complex question is achieved by integrating multiple actions into a single logical form.

\section{Loss Calculation}
\label{sec:loss}

$L^{ed}, L^{el}$, $L^{dec}$ and $L^c$ are the negative log-likelihood losses of the Entity Detection Module, Entity Linking Module, Grammar-guided Decoder and Concept-aware Attention Module, respectively. These losses are defined as follows:
\begin{equation}
    \begin{aligned}
        L^{ed} &= - \sum_{i = 1}^n \log p(y_i^{(ed)} | x), \\
        L^{el} &= - \sum_{i = 1}^n \log p(y_i^{(el)} | x), \\
        L^{dec} &= - \sum_{i = 1}^m \log p(y_k^{(dec)} | x), \\
        L^{c} &= - \sum_{i = 1}^m \log p(y_k^{(c)} | x), \\
    \end{aligned}
\end{equation}
where $n$ and $m$ are the length of the input utterance $x$ and golden logical form, respectively. $ y_i^{(ed)},y_i^{(el)}, y_i^{(dec)},y_i^{(c)}$ are the golden labels for Entity Detection Module, Entity Linking Module, Grammar-guided Decoder and Concept-aware Attention Module, respectively.

\section{Hyper-parameters and Implementation Details}
\label{sec:hyps}
\begin{table}[H]
\centering
\scalebox{1}{
\begin{tabular}{l|c}
\hline
Parameters                  &  Setting \\ \hline
Optimizer                   &  BertAdam\\
Batch Size                  &  120\\
Hidden Size                 &  768\\
Learning Rate               &  5e-5\\
Head Number                 &  6\\
Aggregation Level           &  Token-level\\
Activation Function         &  ReLU\\
Distance  Calculation       &  Absolute\\
Encoder Layer Number        &  2\\
Decoder Layer Number        &  2\\
Loss Component Weight       &  All set to 1\\
GAT Embedding Dimension     &  3072\\
Word Embedding Dimension    &  768\\\hline
\end{tabular}}
\caption{Hyper-parameters for HSGE.}
\label{table:hyper}
\end{table}

The experiments are conducted on 8 NVIDIA V100 GPUs. During model tuning, we identify optimal hyperparameters by modifying one parameter while keeping others fixed and select the hyperparameters that resulted in the highest model performance. We implement our code using Pytorch. The detailed hyper-parameter setting for HSGE is shown in \cref{table:hyper}.

\end{document}